%% file: main.tex
\pdfoutput=1

\documentclass[11pt]{article}

\usepackage[]{acl}

\usepackage{times}
\usepackage{latexsym}

\usepackage[T1]{fontenc}

\usepackage[utf8]{inputenc}

\usepackage{microtype}
\usepackage[whole]{bxcjkjatype}
\usepackage{booktabs}
\usepackage{colortbl}
\usepackage{graphicx}
\usepackage{url}
\usepackage{color}

\title{Target-Guided Open-Domain Conversation Planning}

\author{
Yosuke Kishinami$^{1}$\hspace{1em}
Reina Akama$^{1,2}$\hspace{1em}
Shiki Sato$^{1}$\hspace{1em}
Ryoko Tokuhisa$^{1}$\hspace{1em} \\
\textbf{Jun Suzuki}$^{1,2}$\hspace{1em}
\textbf{Kentaro Inui}$^{1,2}$\hspace{1em} \\
$^{1}$Tohoku University\hspace{1em}
$^{2}$RIKEN\hspace{1em} \\
	\texttt{yosuke.kishinami.q8@dc.tohoku.ac.jp} \\
	\texttt{\{akama,shiki.sato.d1,tokuhisa,jun.suzuki,inui\}@tohoku.ac.jp}
}	

\begin{document}
\maketitle
\begin{abstract}
Prior studies addressing target-oriented conversational tasks lack a crucial notion that has been intensively studied in the context of goal-oriented artificial intelligence agents, namely, \emph{planning}.
In this study, we propose the task of Target-Guided Open-Domain Conversation Planning (TGCP) task to evaluate whether neural conversational agents have goal-oriented conversation planning abilities.
Using the TGCP task, we investigate the conversation planning abilities of existing retrieval models and recent strong generative models.
The experimental results reveal the challenges facing current technology.
\end{abstract}


\section{Introduction}
\label{sec:introduction}
Neural conversational agents have achieved great successes in recent years, and various methods have been proposed to generate informative responses, e.g., the use of knowledge~\cite{Zhao:emnlp2020:KnowledgeGrounded, Wu:acl2020:DiversInformativeDialogue}, personality~\cite{Li:acl2016:PersonabasedNeuralConv,zhang:acl2018:Personachat}, emotional considerations~\cite{Rashkin:acl2019:EmpatheticDialogue, Zhong:emnlp2019:PersonaEmpathetic}, and large-scale models~\cite{Zhang:acl2020:DIALOGPT, Adiwardana:arxiv2020:Meena, Roller:eacl2021:Blender, Thoppilan:arxiv2022:LaMDA}.
One hot topic in this research area is to develop proactive behavior in agents.
For example, \citet{Tang:acl2019:TargetGuided} proposed the task of Target-Guided Open-Domain Conversation, in which an agent is required to actively lead a conversation to a predefined target word.
\citet{Wu:acl2019:ProactiveConv} proposed a task that uses a knowledge graph to actively lead a conversation to a target entity.
Several studies have implemented these target-oriented task settings~\cite{Dai:arxiv2019:MultipleGenerative, Qin:aaai2020:DynamicKnowledge, Yuan:iccai2020:MultihopMemory, Zhong:aaai2021:KeywordGuided, Zhu:sigir2021:ProactiveRetrieval}.
However, these prior studies all lack \emph{planning}, a crucial notion that has been intensively studied in the context of goal-oriented artificial intelligence (AI) agents~\cite[etc.]{NorvigAndRussell:book1995:AIModernApproach, KuijpersAndDockx:ai1998:man-machine-ai-planning, Stent:acl2004:TrainableSentencePlanning, Walker:jair2007:IndividualAndDomain} and has also been introduced in neural conversational agents~\cite{Botea:deep-dial2019:AutomatedPlanning,Jiang:aaai2019:GeneralPlanningBasedFramework,Jiang:sigdial2019:Lookahead}.
In other words, these studies do not explicitly consider the generation of a multiple-step plan to achieve a target.

\begin{figure}[t!]
    \centering
    \includegraphics[width=\columnwidth]{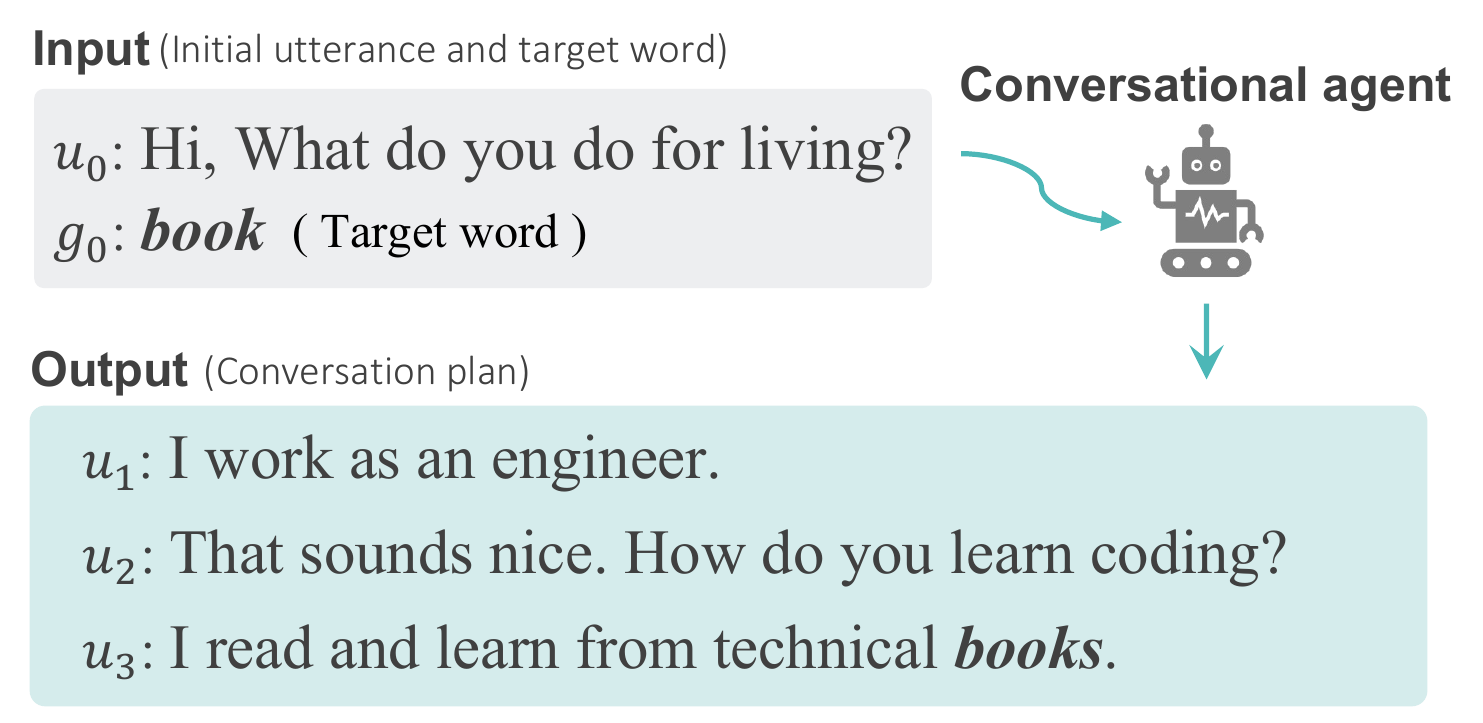}
    \caption{Overview of the TGCP task.}
    \label{fig:task_overview}
\end{figure}

Given this background, in this study, we propose the \textbf{Target-Guided Open-Domain Conversation Planning} (henceforth, \textbf{TGCP}) task such that an agent's planning ability in goal-oriented conversations can be assessed.
The TGCP task is to produce a plan that leads a conversation to a given target, as illustrated in Figure~\ref{fig:task_overview}.
The point is to consider the task of producing a conversation plan for several utterances ahead, which we first address in the aforementioned context of Target-Guided Open-Domain Conversation. 
Furthermore, we also propose modeling the planning process by simulating the user's succeeding utterances using the model of the agent itself; namely, the agent converses with itself (i.e., self-conversation) to search for potential conversation paths that achieve the goal.
This task setting is not the same as a real-world setting, in which an agent is required to plan a conversation while uncertain of the user's future utterances. 
However, planning in the self-conversation setting can be considered a prerequisite capability for a planning-aware goal-oriented conversational agent. 
TGCP works as a framework to evaluate an agents' prerequisite ability for conversation planning without employing human subjects; this can abstract away the hard-to-control human factors from experiments (e.g., some human subjects may not be as cooperative as others).

This paper has three major contributions:
(1) We propose the TGCP task as a framework to assess the prerequisite ability of a model for goal-oriented conversation planning.
(2) We conduct a set of experiments on the TGCP framework using several existing retrieval-based neural models and recently proposed strong generative neural models of conversational agents. 
(3) Our experimental results reveal the challenges facing current technology.
The evaluation codes and the test set used in the experiments are available.\footnote{The evaluation codes and the test set are available at \url{https://github.com/y-kishinami/TGCP}}


\section{Target-Guided Open-Domain Conversation Planning}
We introduce the task of Target-Guided Open-Domain Conversation Planning, the TGCP task for short, that is to evaluate whether the agents have goal-oriented conversation planning abilities.
In this section, we describe the task definition and the evaluation metrics.

\subsection{Task definition}
Figure~\ref{fig:task_overview} shows an overview of the TGCP task.
We define the goal given to the agents as a word (e.g., \textit{dog}, \textit{pizza}, \textit{coffee}).
Given a target word $g_0$ and an initial utterance $u_0$, TGCP requires agents to make an entire conversation plan $(u_1, \dots, u_N)$, whose last utterance $u_N$, which consists of $M$ words, contains the target word $g_0$, namely, $u_N=(w_{N,1}, \dots, w_{N,M})$, and $w_{N,m} = g_0$ for any $m\in M$.
This task has the same input/output format as the human-agent conversation task proposed by \citet{Tang:acl2019:TargetGuided}.
However, these task setups differ in terms of whether or not a human conversational partner is involved.
In TGCP, agents generate for all utterances in the entire conversation.

\subsection{Evaluation metrics}
\label{sec:TGCtask-evaluation}
The evaluation is performed based on three objectives: whether the target word is mentioned (\textbf{achievement ratio}), whether the utterance transitions in the conversation are natural (\textbf{transition smoothness}), and how likely the conversation is to actually occur (\textbf{conversation probability}).
We believe that satisfying these three perspectives is important in goal-oriented conversation planning.
For example, given a target word \textit{computer} and an initial utterance ``What sports do you like?,'' the utterance like \textit{I love computer.} achieves the target, but it is not natural as a conversation, and such an interaction rarely occurs.
Likewise, the utterance like ``\textit{I don't like sports because my friend who likes sports broke my computer.}'' is a natural transition and achieves the target, but it is likely to rarely occur in an actual conversation.
We believe that an agent's generation of such utterances does not indicate the agent's planning ability.
We can automatically calculate the achievement ratio based on whether the target word itself is mentioned or not.\footnote{\citet{Tang:acl2019:TargetGuided} considers mentioning synonyms as the task achievement; however, \citet{Zhong:aaai2021:KeywordGuided} points out that synonyms are unreliable to measure the task achievement. Implementation details are provided in Appendix~\ref{appendix:judgement-achiev}.}
We also consider transition smoothness and conversation probability to be manually evaluated.\footnote{Empirical analyses on the relationship between these two metrics are provided in Appendix~\ref{appendix:analyses-on-relationship}.}

\begin{figure*}[t!]
    \centering
    \includegraphics[width=\textwidth]{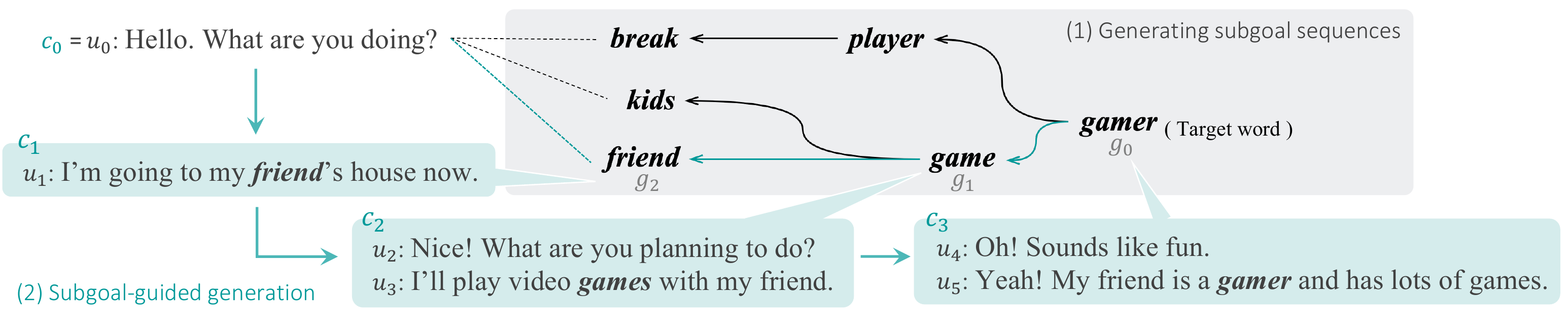}
    \caption{Subgoal-guided conversation plan generation with \textsc{Blener+PreDes.}.}
    \label{fig:predesigning-subgoal}
\end{figure*}


\section{Experiments}
\label{sec:experiment}
Using the proposed TGCP, we investigate the conversation planning ability of several major existing dialogue models and recent deep neural network (DNN) based dialogue models.

\subsection{TGCP settings}
\paragraph{Dataset.}
\label{subsec:exp-dataset}
As a dataset for the TGCP task, we prepared $1,000$ pairs consisting of an initial utterance and a target word, i.e., ($u_0$, $g_0$). 
We created these pairs by randomly extracting from a set of the first utterances and a set of keywords extracted from subsequent utterances in the ConvAI2 dataset.\footnote{This follows existing analogous work~\cite{Tang:acl2019:TargetGuided, Qin:aaai2020:DynamicKnowledge, Zhong:aaai2021:KeywordGuided}. In addition, we removed the keywords not covered by ConceptNet.}

\paragraph{Evaluation.}
As described in Section~\ref{sec:TGCtask-evaluation}, in TGCP, the conversation plans generated by models are evaluated by target achievement ratio, transition smoothness, and conversation probability. The target achievement ratio was calculated automatically. 
To avoid infinite conversations that never reached the target, we set the maximum number of turns to $8$.%
    \footnote{The same setting as previous studies ~\cite{Tang:acl2019:TargetGuided, Qin:aaai2020:DynamicKnowledge, Zhong:aaai2021:KeywordGuided}.}
For transition smoothness and conversation probability, we evaluated them manually using Amazon Mechanical Turk.%
    \footnote{\url{https://www.mturk.com/}}
For each model, randomly sampled $100$ conversation plans were rated by native English speakers. 
We eliminated low-quality workers using attention checks.
Five workers rated each conversation on a five-point Likert scale for transition smoothness ($5$ is \textit{Strongly good} and $1$ is \textit{Strongly bad}) and conversation probability ($5$ is \textit{Frequently} and $1$ is \textit{Rarely}).%
    \footnote{Concrete instructions are provided in Appendix~\ref{appendix:detail-evaluation}.}

\subsection{Existing models}
We prepared the following seven existing dialogue models employed in Target-Guided Open-Domain Conversation:
\citet{Wu:acl2017:SeqMatchNetwork}'s \textsc{Retrieval}, \citet{Tang:acl2019:TargetGuided}'s \textsc{Retrieval-St.}, \textsc{PMI}, \textsc{Neural}, and \textsc{Kernel}, \citet{Qin:aaai2020:DynamicKnowledge}'s \textsc{DKRN}, and \cite{Zhong:aaai2021:KeywordGuided}'s \textsc{CKC}.%
    \footnote{Implementations details are given in Appendix~\ref{appendix:imple-existing-methods}.}
All models except \textsc{Retrieval} are retrieval dialogue models that infer the keyword to mention immediately after each turn of the conversation on the fly and then determine the next response based on it and the conversational history.
\textsc{Retrieval} is a retrieval dialogue model that does not infer the keyword but determine the next response only based on conversational history.

\subsection{Recent DNN-based models}
In addition, we prepared the latest generative model that combines the DNN-based powerful dialogue model, \textsc{Blender}~\cite{Roller:eacl2021:Blender}, and a novel strategy for pre-designing keyword sequences to given $g_0$ (\textsc{PreDes.}).
Our \textsc{Blender+PreDes.} (Figure~\ref{fig:predesigning-subgoal}) is a newly designed model.
Note that this model is new because the task is new, and that there should be many ways to design models for TGCP.
Still, we believe that testing the performance of a specific model such as \textsc{Blender+PreDes.} on TGCP can help investigate the nature of TGCP.
In our \textsc{Blender+PreDes.}, we first generated the keyword sequences, hereinafter it called \textit{subgoal sequence}, using ConceptNet5~\cite{Speer:aaai2017:ConceptNet}. 
Specifically, we acquired the series of $n$ concepts that are passed when tracing the edges of the knowledge graph from the concept representing the target word $g_0$ to the concept related to the initial utterance $u_0$ as a subgoal sequence $G = g_0, g_1, \cdots, g_{n-1}$. $n$ is the length of the subgoal sequence including the target word $g_0$.
This allows preventing cases that cannot get closer to the target than a certain point because of selecting a locally optimal solution.
After generating the subgoal sequence, we generated a sequence of partial conversations $C = c_0, c_1, \cdots, c_{n}$ using \textsc{Blender} as follows:
\begin{equation}
    c_i = f(g_{n-i}, (c_{0}, \cdots, c_{i-1}))\quad(1 \leq i \leq n)
\end{equation}
Where, $c_i$ denotes a partial conversation that follows the previous partial conversation $c_{i-1}$ and ends up with the utterance where the subgoal $g_{n-i}$ appears.
$f(\cdot)$ is a function that returns a partial conversation to the given previous conversations and a subgoal. 

\input{tables/tab-results-all}

\paragraph{\textsc{PreDes.} settings.}
We set $n=3$, i.e., we generated subgoal sequences by tracing ConceptNet up to three levels.%
\footnote{We excluded all stopwords in the NLTK and spaCy libraries to comprehensively exclude unnecessary words. In addition, we excluded the concepts for which the score calculated by wordfreq~\cite{Speer:none:wordfreq} was lower than the score of the target word.}${}^{,}$%
\footnote{We empirically confirmed the validity of this setting (Appendix~\ref{appendix:length-of-subgoal-seq}).}
Among the subgoal sequences generated from ConceptNet, we retained the $30$ subgoal sequences in which the end of the subgoal sequence $g_{n-1}$ was the most related to the given first utterance $u_0$. We calculated the relatedness as the cosine similarity between the SIF embedding~\cite{Arora:iclr2017:SimplebutToughtoBeat} of $u_0$ and GloVe word vector~\cite{Pennington:emnlp2014:GloVe} of $g_{n-1}$.
Among the conversation plans generated from the $30$ subgoal sequences, we selected the conversation plan with the highest average probability of generating partial conversations by \textsc{Blender} as the final output.

\paragraph{Training of \textsc{Blender}.}
We used the Blenderbot $3$B implemented by huggingface transformer as \textsc{Blender}.%
    \footnote{\url{https://github.com/huggingface/transformers}}
Because \textsc{Blender} is a model that generates a response based on the conversational history, we fine-tuned it to use as $f$ which generates a partial conversation based on previous partial conversations and a subgoal.
we used the ConvAI2 processed by \citet{Zhong:aaai2021:KeywordGuided} as a training data for \textsc{Blender}.
We prepared the training data by randomly splitting a single conversation into an input and an output consisting of multiple utterances and then concatenating a word extracted randomly from the output-side utterances (i.e., keywords) to the input utterances.%
\footnote{We extracted the keywords by following \citet{Zhong:aaai2021:KeywordGuided}.
The pairs that failed to extract keywords from the output utterances were excluded from the training data.}
We finally obtained $117,\!877$ pairs as training set and $6,\!425$ pairs as validation set.
The hyperparameters are provided in Appendix~\ref{appendix:training-details}.

\subsection{Ablation models}
To analyze the effectiveness of the pre-design strategy, we also prepared Blender without any conversational strategy (\textsc{Blender}), and with an on-the-fly strategy using existing models.
Specifically, we employed the strategy of CKC as the comparison on-the-fly strategy, which is known to be the highest performance method in TGC~\cite{Zhong:aaai2021:KeywordGuided} (\textsc{Blender+CKC}).
For both models, Blender is the same as \textsc{Blender+PreDes.}.
But note that blender, without any conversation strategy, does not concatenate keywords with inputs for training and inference.

\subsection{Results}
\label{subsec:experiments2-results}
Table \ref{tab:results-all} shows the evaluation results on TGCP.
To provide the human upper bound performance, we also had three workers perform TGCP on $50$ pairs randomly selected from the dataset described in Section~\ref{subsec:exp-dataset} (Human).

\paragraph{Achievement ratio.}
The achievement ratios of the retrieval models tended to be high.
In particular, the achievement ratio of \textsc{DKRN} was comparable to that of humans.
The generative models had lower achievement ratios.
However, \textsc{Blender$+$PreDes.} improved the achievement ratio compared with \textsc{Blender+CKC}, whose subgoal strategy is the same as that of \textsc{CKC}.
This result means that replacing the on-the-fly subgoal strategy with the pre-design strategy is effective in improving the target achievement ratios of generative models.

\paragraph{Smoothness \& probability.}
The retrieval models have lower values of transition smoothness and conversation probability than humans.
Table~\ref{tab:DKRN-example} shows a conversation plan example generated by \textsc{DKRN}, whose achievement ratio was the highest of all the compared models.
In the example, the transition between $u_{1}$ and $u_{2}$ is clearly unnatural, although the model achieved to mention the target word.%
    \footnote{An additional example is provided in Appendix~\ref{appendix:deadend-example}.}
The transition smoothness and conversation probability of the generative models were higher than those of the retrieval models.
In particular, \textsc{Blender+CKC} significantly outperformed \textsc{CKC} in these metrics.
Therefore, using powerful DNN-based generation models improves the transition smoothness and conversation probability of the conversation plans.
Table~\ref{tab:predes-example} shows a conversation plan example generated by \textsc{Blender$+$PreDes.}, whose transition smoothness was the highest of all the compared models.
In this example, \textsc{Blender$+$PreDes.} generated a natural conversation along an appropriately generated subgoal sequence.

\paragraph{Overall.}
The TGCP task revealed the planning abilities of well-known retrieval models and newly prepared generative models.
The retrieval models tended to have high achievement ratios but low transition smoothness and conversation probability, while the opposite was true for the generative models.
These results show the trade-off between achievement ratio and the naturalness of conversation plans that current technology is facing.
On the other hand, the generative model with a pre-design subgoal strategy (\textsc{Blender$+$PreDes.}) improved the achievement ratio compared with the generative model with an on-the-fly strategy (\textsc{Blender$+$CKC}) ensuring its high transition smoothness and conversation probability.
This implies that improving the achievement ratios of generative models by refining their subgoal strategies is an effective approach to overcome the trade-off.

\subsection{Discussion: Number of conversation turns}
We found that generative models behave critically differently from humans regarding the number of turns to reach targets, while their transition smoothness and conversation probability were comparable to those of humans.
The average numbers of turns to reach targets of \textsc{Blender$+$CKC} and \textsc{Blender$+$PreDes.} were much larger than that of humans.%
\footnote{We counted the number of turns of the conversations where the target words are mentioned.}
This result indicates that humans efficiently achieved TGCP with fewer turns while ensuring high conversation probability.%
    \footnote{A generated example is provided in Appendix~\ref{appendix:conv-example}.}
Making an agent has a strong conversation planning ability like a human can be one of our challenges in the future.

\input{tables/tab-conversation-plan-dkrn}


\section{Conclusion}
We have proposed the TGCP task as a platform for assessing the conversation planning ability of a dialogue model.
Through this task setting, we have presented a first study for assessing the present neural conversational models' abilities for multiple-utterance planning, abstracting away the hard-to-control potential human factors. 
While the reported experiments cover only the task of Target-Guided Open-Domain Conversation~\cite{Tang:acl2019:TargetGuided}, the idea of TGCP is expected to be applicable to a wider range of goal-oriented conversation tasks.
Using TGCP, we revealed that the dialogue models with current technology have difficulty planning conversations to achieve given goals while ensuring the naturalness of the conversation.
The experimental results also showed that refining the subgoal strategies for generative models might be an effective method to overcome this trade-off.
We plan to research methods to solve this task setting with higher performance.

\input{tables/tab-conversation-plan-blender+predes}


\section*{Acknowledgments}
We would like to thank all anonymous reviewers for their invaluable comments.
This work was partly supported by JSPS KAKENHI Grant Numbers JP21J22383, JP22K17943, JST Moonshot R\&D Grant Number JPMJMS2011.


\clearpage
\section*{Ethical considerations}
This paper honors the ACL Code of Ethics. 
This study uses existing dataset, preprocessed versions of ConvAI dataset~\citet{Tang:acl2019:TargetGuided, Zhong:aaai2021:KeywordGuided}, which we believe does not involve any ethical concerns.
We used these data as training data for dialogue models and keyword prediction models and to randomly extract subsets for use as inputs for our generation task; we do not believe this involves any ethical concerns.
This study includes manual work; For human evaluation, we hired crowdworkers and paid appropriate rewards for their labor (corresponding to a \$$14.40$ for an hourly wage).


\bibliography{references}
\bibliographystyle{acl_natbib}

\clearpage
\appendix


\section{Details of the TGCP Task}
\subsection{Calculation of Target Achievement Ratio}
\label{appendix:judgement-achiev}
The achievement judgment was based on whether the target word itself was mentioned~\cite{Zhong:aaai2021:KeywordGuided}; however, we found that there were several cases in which the achievement was judged to be a failure even though the target word was appropriately mentioned. Therefore, we modified the script\footnote{\url{https://github.com/zhongpeixiang/CKC/blob/master/util/data.py}} for judging the achievement such that these cases would be judged as achievements.

\subsection{Relationship between Transition Smoothness and Conversation Probability}
\label{appendix:analyses-on-relationship}
We investigated the correlation between transition smoothness and conversation probability of the experiment in Section~\ref{sec:experiment}, and found that Pearson's correlation coefficient was $0.828$, which indicates a high correlation.
Therefore, it appears that conversation probability is contained within transition smoothness, at least in our experiment.
Therefore, evaluating transition smoothness may indicate the approximate tendency of conversation probability.


\section{Model Implementations and Setups}
\label{appendix:model-imple-and-setup}
\subsection{Existing Models}
\label{appendix:imple-existing-methods}
We used publicly available codes by their authors to implement the existing models.%
    \footnote{\url{https://github.com/James-Yip/TGODC-DKRN}}${}^{,}$%
    \footnote{\url{https://github.com/zhongpeixiang/CKC}}
To train the response selection models and the keyword prediction models, we used the same dataset and setups as described in their papers: \textsc{CKC} used the ConvAI2 dataset processed by \citet{Zhong:aaai2021:KeywordGuided}, and the other models used the ConvAI2 dataset processed by \citet{Tang:acl2019:TargetGuided}.

\subsection{Training parameters of \textsc{Blender}.}
\label{appendix:training-details}
To train \textsc{Blender}, we set the batch size to $32$, the learning rate to $7.0\times 10^{-6}$, the warmup steps to $100$, the evaluation steps to $1,\!000$, and the number of updates to $50,\!000$.  The other parameters were set to the default configuration of the huggingface transformer.
We used the model at the validation loss minimum point for conversation planning.

\subsection{Length of subgoal sequence.}
\label{appendix:length-of-subgoal-seq}
We empirically confirmed the validity of tracing ConceptNet up to the three levels using the following procedure.
First, we qualitatively checked the subgoal sequences generated by \textsc{PreDes.} and found that the subgoal sequences with relatedness scores of approximately $0.6$, indicating a connection with the initial utterance, were naturally connected with the initial utterance.
Then, we investigated how much ConceptNet need to be searched  to generate subgoal sequences with a score of approximately $0.6$.
As a result, we confirmed that by tracing ConceptNet up to three steps, the average score of the subgoal sequences of the search results was $0.653$, which exceeded $0.6$.
Therefore, we conclude that the three-step search is reasonable.


\section{Instructions for Human Evaluation}
\label{appendix:detail-evaluation}
\begin{figure*}[t]
    \centering
    \includegraphics[width=\textwidth]{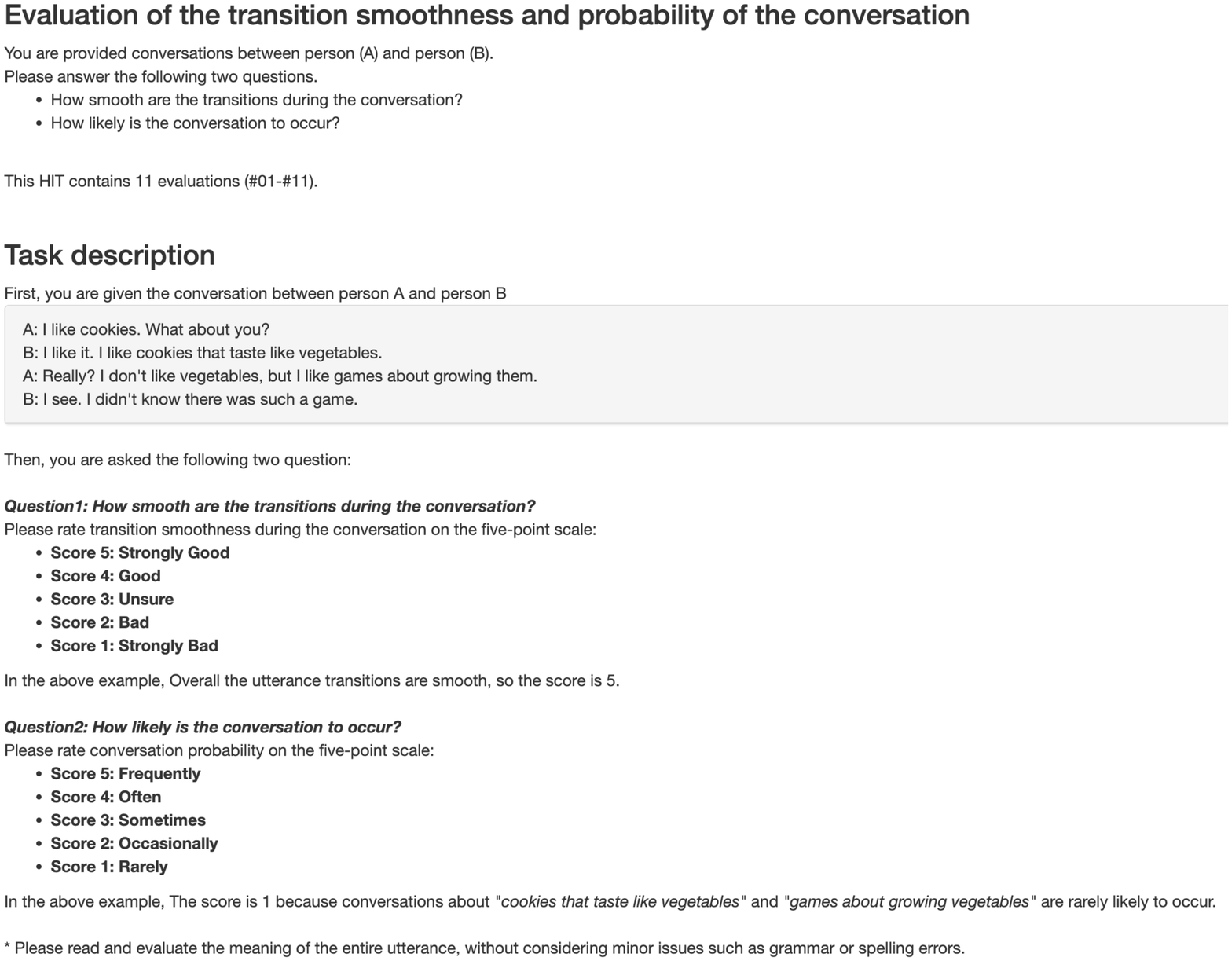}
    \caption{Evaluation instructions given to Amazon Mechanical Turk workers.}
    \label{fig:amt-instcution}
\end{figure*}

Figure~\ref{fig:amt-instcution} shows the instructions given to Amazon Mechanical Turk workers concerning the evaluation of transition smoothness and conversation probability.


\section{Generated Conversation Plans}
\label{appendix:additional-examples}
\subsection{Human-generated Conversation Plan}
\label{appendix:conv-example}
Table~\ref{tab:human-example} shows a conversation plan generated by a human.
We confirmed that human could plan natural conversations that achieved their target despite the short number of turns.

\begin{table}[t]
\centering
\small
\tabcolsep 3pt
\renewcommand{\arraystretch}{1.5}
\begin{tabular}{cp{68mm}}
\toprule
$u_0$ & Not a big fan of talking face to face . How about you? \\
\rowcolor{gray!7}
$u_1$ & Me too. I prefer texting. \\
$u_2$ & I truly understand. People love to comment on your behaviors when talking face to face. But talking online does not have such problems. \\
\rowcolor{gray!7}
$u_3$ & It sounds like you have experienced such comments. What did people accuse you of? \\
$u_4$ & Well, I am a \textit{\textbf{vegetarian}}, but they said vegetarians are incomprehensible. Rude people, aren't they? \\
\bottomrule
\end{tabular}
\caption{Conversation plan generated by a human. (Target: \textit{\textbf{vegetarian}})}
\label{tab:human-example}
\end{table}

\subsection{Dead-ended Case}
\label{appendix:deadend-example}

Table \ref{tab:retrieval-stgy-example} shows an example where \textsc{Retrieval-St.} did not achieve its target.
After the keyword \textit{catch} is selected in utterance $u_6$, \textsc{Retrieval-St.} continued to generate the same utterance repeatedly from $u_7$ to the end ($u_{16}$).
Ultimately, \textsc{Retrieval-St.} could not reach the target.

\input{tables/tab-conversation-plan-retrieval-st}

\end{document}

%% file: tables/tab-results-all.tex
\begin{table*}[t]
    \centering
    \small
    \tabcolsep 5pt
    \begin{tabular}{l cc l cccc}
        \toprule
        Model & Subgoal & Conversation && Achievement & \#Turns & Smoothness & Probability \\
        \cmidrule{1-3} \cmidrule{5-8}
        \textsc{Retrieval}~\cite{Wu:acl2017:SeqMatchNetwork} & - & retrieval
        && 0.034 & 3.71 & 3.52 & 3.37 \\
        \textsc{Retrieval-St.}~\cite{Tang:acl2019:TargetGuided} & on-the-fly & retrieval
        && 0.851 & 5.04 & 3.50 & 3.29 \\
        \textsc{PMI}~\cite{Tang:acl2019:TargetGuided} & on-the-fly & retrieval
        && 0.531 & 4.97 & 3.33 & 3.17 \\
        \textsc{Neural}~\cite{Tang:acl2019:TargetGuided} & on-the-fly & retrieval
        && 0.535 & 2.83 & 3.14 & 3.00 \\
        \textsc{Kernel}~\cite{Tang:acl2019:TargetGuided} & on-the-fly & retrieval
        && 0.596 & 2.79 & 3.24 & 3.05 \\
        \textsc{DKRN}~\cite{Qin:aaai2020:DynamicKnowledge} & on-the-fly & retrieval
        && \textbf{0.968} & 2.91 & 3.28 & 3.12 \\
        \textsc{CKC}~\cite{Zhong:aaai2021:KeywordGuided}& on-the-fly & retrieval
        && 0.353 & 3.60 & 2.81 & 2.69 \\
        \midrule
        \textsc{Blender}~\cite{Roller:eacl2021:Blender} & - & generative
        && 0.024 & 5.04 & 3.99 & \textbf{3.90} \\
        \textsc{Blender$+$CKC} & on-the-fly & generative
        && 0.247 & 7.00 & 3.90 & 3.71 \\ 
        \textsc{Blender$+$PreDes.} & pre-design & generative
        && 0.425 & 6.29 & \textbf{4.05} & \textbf{3.90} \\
        \midrule
        Human & - & - && 1.000 & 3.50 & 4.11 & 3.89 \\
        \bottomrule
    \end{tabular}
    \caption{Performance of dialogue models on the TGCP task.}
    \label{tab:results-all}
\end{table*}

%% file: tables/tab-conversation-plan-dkrn.tex
\begin{table}[t!]
    \centering
    \small
    \tabcolsep 3pt
    \renewcommand{\arraystretch}{1.5}
    \begin{tabular}{cp{64mm}}
        \toprule
        $u_0$ & hey how is it going ? \\
        \rowcolor{gray!7}
        $u_1$ & i'm doing ok . i have mass this week (\textit{school}: 0.64) \\
        $u_2$ & i just got done sewing a new shirt (\textit{shirt}: 1.00) \\
        \bottomrule
    \end{tabular}
    \caption{Part of the conversation plan by an existing model (\textsc{DKRN}). The elements in parentheses are keywords predicted by the model and the similarity score between the keyword and the target. (Target: \textit{\textbf{shirt}})}
    \label{tab:DKRN-example}
    \vspace{-1mm}
\end{table}

%% file: tables/tab-conversation-plan-blender+predes.tex
\begin{table}[t!]
    \centering
    \small
    \tabcolsep 3pt
    \renewcommand{\arraystretch}{1.5}
    \begin{tabular}{cp{68mm}}
        \toprule
        $u_0$ & hello how are you ? \\
        \rowcolor{gray!7}
        $u_1$ & not too bad, how about you? \\
        $u_2$ & i am good thanks for asking \\
        \rowcolor{gray!7}
        $u_3$ & what do you do for a living, if you don't mind me asking? i am a nurse \\
        $u_4$ & that's cool. i work in a grocery store. \\
        \rowcolor{gray!7}
        $u_5$ & do you like it \\
        $u_6$ & yeah, it pays the bills, but i want to be a dental hygienist \\
        \rowcolor{gray!7}
        $u_7$ & wow that's a great career choice. how long have you been doing that \\
        $u_8$ & for as long as i can \textit{remember} \\
        \rowcolor{gray!7}
        $u_9$ & what do you like to do in your spare time \\
        $u_{10}$ & i love to take \textit{pictures} and photography is a hobby of mine \\
        \rowcolor{gray!7}
        $u_{11}$ & what kind of pictures do you take? \\
        $u_{12}$ & mostly \textit{landscapes}, i love nature \\
        \bottomrule
    \end{tabular}
    \caption{Conversation plan by \textsc{Blender$+$PreDes.} (Target: \textit{\textbf{landscape}}). The predicted subgoal sequence is \textit{remember} $\rightarrow$ \textit{picture} $\rightarrow$ \textit{\textbf{landscape}}.}
    \label{tab:predes-example}
    \vspace{-1mm}
\end{table}

%% file: tables/tab-conversation-plan-retrieval-st.tex
\begin{table}[t]
    \centering
    \small
    \tabcolsep 3pt
    \renewcommand{\arraystretch}{1.5}
    \begin{tabular}{cp{64mm}}
        \toprule
        $u_0$ & Today I am annoyed about not being able to drive. \\
        \rowcolor{gray!7}
        $u_1$ & That's \textit{horrible}. My friend is also paralyzed. (\textit{horrible}: 0.20) \\
        $u_2$ & I am sorry to \textit{hear} of your injury. (\textit{hear}: 0.43) \\
        \rowcolor{gray!7}
        $u_3$ & I feel like dump and skeleton but just \textit{wasting} my hours here. (\textit{waste}: 0.48) \\
        $u_4$ & I don't understand \textit{people} with egos. (\textit{people}: 0.50) \\
        \rowcolor{gray!7}
        $u_5$ & I think happiness is an illusion which \textit{money} can help with. (\textit{money}: 0.53) \\
        $u_6$ & I don lie, its bad when you get \textit{caught}. (\textit{catch}: 0.54) \\
        \rowcolor{gray!7}
        $u_7$ & I guess that's why the constantly keep calling me. (\textit{catch}: 0.54) \\
        $u_8$ & I guess that's why the constantly keep calling me. (\textit{catch}: 0.54) \\
        \rowcolor{gray!7}
        $\vdots$ & \\
        $u_{16}$ & I guess that's why the constantly keep calling me. (\textit{catch}: 0.54) \\
        \bottomrule
    \end{tabular}
    \caption{Part of the conversation plan by an existing model (\textsc{Retrieval-St.}). The elements in parentheses are keywords predicted by the model and similarity score between the keyword and target. (Target: \textit{\textbf{chase}})}
    \label{tab:retrieval-stgy-example}
\end{table}